\documentclass[a4paper, runningheads, 11pt]{llncs}

\usepackage{amsmath}
\usepackage{amsfonts}
\usepackage{amssymb}
\usepackage{graphicx}
\usepackage{extpfeil}

\numberwithin{equation}{section}
\begin{document}
\title{On the Conditions for Domain Stability for Machine Learning: a Mathematical Approach}
\author{Gabriel Pedroza\\ \orcidID{0000-0002-7889-2892}}
\authorrunning{Gabriel Pedroza}
\titlerunning{Mathematical conditions for Domain Stability}

\institute{Ansys, France\\
\email{gabriel.pedroza@ansys.com, pedrozafm@gmail.com}\\
\url{www.linkedin.com/in/gabriel-pedroza-89bb5338}}

\maketitle

\begin{abstract}
This work proposes a mathematical approach that (re)defines a property of Machine Learning models named stability and determines sufficient conditions to validate it. 
Machine Learning models are represented as functions, and the characteristics in scope depend upon the domain of the function, what allows us to adopt topological and metric spaces theory as a basis. Finally, this work provides some equivalences useful to prove and test stability in Machine Learning models. The results suggest that whenever stability is aligned with the notion of function smoothness, then the stability of Machine Learning models primarily depends upon certain topological, measurable properties of the classification sets within the ML model domain.
\end{abstract}

\section{Introduction}

In recent years, the study of Artificial Intelligence and related techniques, like Machine Learning, has attracted considerable attention from practitioners, engineers and researchers from many sectors. 
Such interest has been, in many respects, driven by questions and concerns raised and shared by the involved communities. In particular, it can be mentioned the need for safety of systems integrating AI algorithms and the possibility to achieve acceptable means for compliance, necessary for regulation and certification \cite{EUAIAct}\cite{DEELProject}\cite{Confiance.AIProject}.
Despite autonomy is a well-known notion within the Systems, SW and HW engineering arenas, the usage of AI technology, (GenAI, LLMs, ML/DL) to carry out functions traditionally conducted under human supervision and control, induces new challenges \cite{FischerAIsystemengineering}.
Discussing referred challenges is out of the scope of this paper: notwithstanding the relevance of AI challenges and risks, they are currently under inspection and study by a variety of experts, following a plethora of approaches, ranging from conceptual to empirical/pragmatical \cite{Habbal2024}\cite{piorkowski2022quantitative}. 
From the variety and heterogeneity of those studies, a consensus seems to emerge: despite the significant advances in AI technology, a considerable amount of requirements and properties still need to be ensured for the AI systems to be trustworthy and aligned with requirements.
It is known that the lack of properties like robustness, generalization, and stability in ML models can have impacts at different system levels \cite{RobustnessAI2024}. Several authors even highlight a lack of methods to validate and ensure referred properties \cite{FoundationsAIHarel2020}. In addition, the observed uncertainty of ML performance raise questions on the foundations of the AI algorithms. A basis for their sound design and validation seems necessary but still missing. This work aims to provide a preliminary mathematical basis to analyze ML models abstracted as functions.
It mainly introduces a property named stability, which appears to be fundamental for ML classification. This preliminary work provides some equivalences which can be useful to prove stability. Topological and metric spaces \cite{MeasureTheoryHalmos}\cite{TopologicalManifoldsJohnLee} were adopted as foundation to conduct this work which also relies on the theory of functional analysis \cite{FunctionalAnalysisLang}. 

\section{Defining Stability}

\begin{definition}\label{def:classifier}
Let $(S, d)$ a metric space, $D_{i}\subset S$, $i=1 \ldots m$, a finite sequence of sets. $M$ is a classifier on $S$ for the sets $\{D_{i}\}$ if:
\begin{itemize}
    \item [i)] $M$ is a function defined on $\bigcup_{i=1}^{m} D_{i}$
    \item [ii)] $D_{i} \cap D_{j} = \emptyset$, $\forall i \neq j$
    \item [iii)] $\exists y_{i} \in M(D_{i})$, $\forall x\in D_{i} \Rightarrow M(x)=y_{i},\ \ i=1 \ldots m$
    \item [iv)] For $y_{i},\ y_{j}$ as in previous point, if $i \neq j\ \Rightarrow\ y_{i} \neq y_{j}$
\end{itemize}
\end{definition}

\begin{note}\label{not:neighborhood}
 Given a metric space $(S, d)$ the following notation is used for the neighborhood of a point $x_{o} \in S$:
 \begin{center}
 $B(x_{o},\delta)=\{ x \in S\ |\ d(x_{o},x)<\delta\}$
 \end{center}
 \end{note}

\begin{note}\label{not:domainpoints}
    In Definition \ref{def:classifier}, a point $x\in D_{i}$, $M(x)=y_{i}$ is denoted by $x_{y_{i}}$. The assignation $M(x_{y_{i}})=y_{i}$ can also be denoted by $x_{y_{i}} \xmapsto {M} y_{i}$.
\end{note}

\begin{definition}\label{def:stablepoints}
Let $(S, d)$ a metric space, $D\subset S$ and $M$ a classifier for $D$ and $D^{c}$. $x_{y} \in D$ is said a stable point of $M$ in $D$ if it is satisfied:
\begin{itemize}
    \item [i)] $M(x_{y})=y$ 
    \item [ii)] $\exists \delta > 0,\ \forall x \in B(x_{y},\delta)$ $\Rightarrow$ $x \in D$, $M(x)=y$
    \item [iii)] For $\delta$ as in point $ii)$, $\forall \delta_{\alpha} \leq \delta$ $\exists x \in B(x_{y}, \delta_{\alpha})$, $x \neq x_{y}$ 
\end{itemize}
\end{definition}

Practically, the existence of stable points for a classifier $M$ implemented in a programming language, relies upon the limited/finite precision of computers. Indeed, the precision of a machine can be approximated, for instance, by an iterative algorithm taking $\varepsilon_{o} > 0$ as input and computing $\varepsilon_{n+1} = \varepsilon_{n}/2^{n}$ at each iteration $n$. The stopping criterion is when $1+\varepsilon_{n} = 1$. Then, a candidate for $\delta$ is any number $k-$times bigger than $\varepsilon_{n}$. i.e. $\delta:=k \varepsilon_{n}$.\\
The Definition \ref{def:stablepoints} helps to identify cases where classifiers are unable to smoothly classify subsets within its domain, as can be seen in the following example.

\begin{example}
Let $M$ be a classifier in the interval $S=[0,1]$, with the 1-euclidean metrics for the sets $D_{1}=\mathbb{I} \cap S$ and $D_{2}=\mathbb{Q} \cap S$. Given that $D_{1}$ and $D_{2}$ are dense, then the set of stable points of $M$ in $D_{i}$ is empty.  
\end{example}

\section{Dense sets and Implications for Stability}

The implications of dense sets regarding the existence (absence) of stable points is formalized in this subsection.

\begin{definition}
Let $(S, d)$ a metric space. A set $D \subset S$ is said dense in $S$ if $\forall x \in S$, $\forall \delta >0$, $B(x,\delta) \cap D \neq \emptyset$ \cite{TopologicalManifoldsJohnLee}, \cite{MeasureTheoryHalmos}.
\end{definition}

\begin{lemma}\label{lemma:densesets}
Let $(S,d)$ a metric space and $M$ a classifier for $D \subset S$ and $D^{c}$. If $D^{c}$ is dense in $S$ then the set of stable points in $D$ is empty.
\end{lemma}
First, $S=D \cup D^{c}$. Let's assume $x_{o} \in D$ a stable point of $M$, then $\exists \delta >0$ such that $B(x_{o},\delta) \subset D$. Since $D^{c}$ is dense in $S$, then $B(x_{o},\delta) \cap D^{c}\neq \emptyset$. However, if $x\in B(x_{o},\delta) \cap D^{c}$ then $x\in D^{c}$ and by definition \ref{def:stablepoints}, $ii)$ $x\in D$ what contradicts the assumption. QED.\\

By exchanging roles between $D$ and $D^{c}$ in previous lemma, the following corollary is proved.

\begin{corollary}\label{corollary:densesets}
Let $(S,d)$ a metric space and $M$ a classifier for $D \subset S$ and $D^{c}$. If $D$ is dense, then $D^{c}$ does not have any stable point. 
\end{corollary}

\begin{corollary}
Let $S \subset \mathbb{R}$ a (non-empty) interval, then there is no classifier $M$ able to classify $S \cap \mathbb{I}$ and $S \cap \mathbb{Q}$. 
\end{corollary}
Indeed, since $(S \cap \mathbb{Q}) = (S \cap \mathbb{I})^{c}$ and $S \cap \mathbb{I} = (S \cap \mathbb{Q})^{c}$ are both dense, by lemma \ref{lemma:densesets} and Corollary \ref{corollary:densesets} the conclusion follows. QED.\\

The previous results, which hold particularly for intervals in $\mathbb{R}$, are generalized in the following lemma.

\begin{lemma}\label{lemma:genClassifierDenseSets}
Let $(S,d)$ a metric space and $M$ a classifier for a collection of sets $D_{1} \ldots D_{n} \subset S$ such that $S=\cup_{i=1}^{n} D_{i}$. If there is $D_{k}$ dense in $S$ then there is no stable point for $M$ in any of the sets $D_{i}$ for $i\neq k$.  
\end{lemma}
Let's assume there is $x_{j} \in S$ such that $x_{j} \in D_{j},\ j \neq k$, is a stable point for $M$. Then $M(x_{j})=j$ and $\exists \delta>0$ such that $B(x_{j},\delta) \subset D_{j}$. However, since $D_{k}$ is dense in $S$, then $B(x_{j}, \delta) \cap D_{k}\neq \emptyset$. Then for any point $x \in B(x_{j}, \delta) \cap D_{k}$, it occurs that $M(x) = j$ and $M(x) =k$, and $x\in D_{j} \cap D_{k}$ which contradicts Definition \ref{def:classifier} $i)$, $ii)$, $iv)$ and Definition \ref{def:stablepoints}, $ii)$. QED.    

\section{Alternatives to Prove Stability}

Some equivalences are provided as alternatives to prove stability.  

\subsection{Accumulation Points}

\begin{definition}\label{def:accumulationpoint}
    Let $(S,d)$ a metric space. Given a set $D \subset S$, a point $x \in S$ is an accumulation point of $D$ if $\forall \delta >0$ then $B(x,\delta) \cap D \backslash \{ x\} \neq \emptyset$ \cite{TopologicalManifoldsJohnLee} \cite{MeasureTheoryHalmos}.
\end{definition}

\begin{lemma}\label{lemma:eqAccumulationPointStability}
Let $(S,d)$ a metric space and $M$ a classifier for $D$ and $D^{c}$, with both $D$ and $D^{c}$ not dense in $S$ and $D$ an open set. Then $x_{y} \in D$ is a stable point for $M$ if and only if $x_{y}$ is an accumulation point of $D$.
\end{lemma}

[$\mathbf{\Rightarrow}$]\\
Let's assume $x_{y}\in D$ is a stable point of $M$. It will be proved that $x_{y}$ is an accumulation point of $D$.\\ 
Let's consider $\varepsilon >0$ arbitrary but fixed. Then since $x_{y}\in D$ is a stable point of $M$, $M(x_{y})=y$ and $\exists \delta > 0$ such that $B(x_{y},\delta) \subset D$ and $\forall \delta_{\alpha} \leq \delta$, $\exists x \in B(x_{y}, \delta_{\alpha}),\ x \neq x_{y}$. The following cases exist:\\
%Let's assume $\exists \delta_{\alpha}>0$ such that $B(x_{y},\delta_{\alpha})\cap D \backslash \{ x_{y} \} = \emptyset$.\\

If $\varepsilon < \delta$: then $B(x_{y},\varepsilon) \subset B(x_{y},\delta) \subset D$, since  $\exists x \in B(x_{y}, \varepsilon),\ x \neq x_{y}$, it follows that $x \in D$, which leads directly to $B(x_{y},\varepsilon)\cap D \backslash \{ x_{y} \} \neq \emptyset$.\\

If $\varepsilon > \delta$: since $x_{y}\in D$ is stable point of $M$, then $\exists x \in B(x_{y},\delta) \subset B(x_{y},\varepsilon)$,  $x\neq x_{y}$ and $x \in D$, what also leads to $B(x_{y},\varepsilon) \cap D \backslash \{ x_{y} \} \neq \emptyset$.\\

[$\mathbf{\Leftarrow}$]\\
Now, let's assume $x_{y}$ is an accumulation point of $D$, an open set. It will be proved that $x_{y} \in D$ is a stable point of $M$.\\ 

By definition, $\forall \delta >0,\ B(x_{y}, \delta) \cap D \backslash \{ x_{y} \} \neq \emptyset$. Let $\delta > 0$ arbitrary but fixed. Then, since $\forall \delta_{\alpha} \leq \delta$, $B(x_{y},\delta_{\alpha})\cap D \backslash \{x_{y}\} \neq \emptyset$, this implies that $\exists x_{\alpha}\in B(x_{y},\delta_{\alpha})\cap D$, $x_{\alpha} \neq x_{y}$. Therefore, $M(x_{\alpha})=y$ since $x_{\alpha} \in D$ and $M$ classifier for $D$ and $D^{c}$. Previous statement holds for any $\delta_{\alpha} \leq \delta$. Let's assume $\forall \delta_{\alpha} < \delta$, $B(x_{y}, \delta_{\alpha}) \cap D^{c} \neq \emptyset$. Since $D$ is open and $x_{y} \in D$, then $\exists \varepsilon >0$ such that $B(x_{y}, \varepsilon) \subset D$. The following cases exist:\\

If $\varepsilon < \delta$: then $B(x_{y},\varepsilon) \subset D$ and $B(x_{y}, \varepsilon) \cap D^{c} \neq \emptyset$  
what is a contradiction\\

If $\varepsilon > \delta$: then $B(x_{y},\delta) \subset B(x_{y},\varepsilon) \subset D$, what conflicts with the assumption $\forall \delta_{\alpha} < \delta$, $B(x_{y}, \delta_{\alpha}) \cap D^{c} \neq \emptyset$\\

Therefore $\exists \delta_{\alpha} < \delta$ such that $B(x_{y}, \delta_{\alpha}) \cap D^{c} = \emptyset$ what implies $B(x_{y}, \delta_{\alpha}) \subset D$, thus $\forall x \in B(x_{y}, \delta_{\alpha})\ \Rightarrow M(x) = y$. QED.\\

\subsection{Accumulation series}

As shown in previous subsection, a classifier $M$ defined over an open set is stable for every accumulation point therein. The equivalence provided in the following lemma provides further means to prove stability.

\begin{lemma}\label{lemma:eqAccumPointsAndSeries}
Let $(S,d)$ a metric space and $M$ a classifier for $D$ and $D^{c}$, with both $D$ and $D^{c}$ not dense in $S$ and $D$ an open set. $x_{y}$ is a stable point of M if and only if $\forall \{x_{n}\}$ series, such that $lim_{n \to \infty} x_{n} = x_{y}$, $x_{n} \neq x_{y}$, $\exists \{ s_{k}\}$ sub-series of $\{x_{n}\}$ such that $\exists k_{o},\ k \geq k_{o}$, $\Rightarrow$ $s_{k} \in D,\ s_{k} \neq x_{y}$, $lim_{k \to \infty} s_{k} = x_{y}$.    
\end{lemma}

[$\mathbf{\Rightarrow}$]\\
Let's assume $x_{y}$ is a stable point of $M$. Let $\{ x_{n}\}$ be a series such that $lim_{n \to \infty} x_{n} = x_{y}$, $x_{n} \neq x_{y}$, then $\exists \delta >0$ such that $B(x_{y},\delta) \subset D$. For such $\delta$, $\exists n_{o}$ such that $\forall n > n_{o}$, $d(x_{y},x_{n})<\delta$, what means that $\forall n > n_{o},\ x_{n} \in B(x_{y},\delta) \subset D$, then $M(x_{n})=y$, $x_{n} \neq x_{y}$. Then, we can define the sub-series of $\{ x_{n} \}$ as  $\{ s_{k}\} = \{ x_{n}\} \cap B(x_{y},\delta) \cap D \backslash \{ x_{y}\} \neq \emptyset$. Thus, by its construction, $\{ s_{k}\}$ satisfies $\exists k_{o},\ k \geq k_{o}$, $\Rightarrow$ $s_{k} \in D,\ s_{k} \neq x_{y}$, $lim_{k \to \infty} s_{k} = x_{y}$.\\

[$\mathbf{\Leftarrow}$]\\
%Let's assume $\forall \{x_{n}\}$ series, such that $lim_{n \to \infty} x_{n} = x_{y}$, $\exists \{ s_{k}\}$ a sub-series of $\{x_{nk\}$ such thats$\kxists k_{o},\ k \geq k_{o}$, $\Rightarrow$ $s_{k} \in D,\ s_{k} \neq x_{y}$. It will be proved that $x_{y}$ is an accumulation point of $D$, open set.\\
Let $\delta >0$. Then, if $lim_{n \to \infty} x_{n} = x_{y}$, $x_{n} \neq x_{y}$, and $\exists \{ s_{k}\}$ a sub-series of $\{x_{n}\}$ such that $\exists k_{o},\ k \geq k_{o}$, $\Rightarrow$ $s_{k} \in D,\ s_{k} \neq x_{y}$, $lim_{k \to \infty} s_{k} = x_{y}$, let $\varepsilon = d(s_{k_{o}}, x_{y})$. The following cases exist:\\

If $\varepsilon < \delta$: $\forall k > k_{o}$, $\Rightarrow$ $s_{k} \in B(x_{y}, \varepsilon) \subset B(x_{y}, \delta)$ and $s_{k} \in D,\ s_{k} \neq x_{y}$, therefore $\forall k > k_{o}$, $s_{k} \in D \cap B(x_{y}, \delta) \backslash \{ x_{y} \} \neq \emptyset$.\\

If $\varepsilon > \delta$: since $lim_{k \to \infty} s_{k} = x_{y}$, then for the given $\delta$, $\exists k_{1}$ such that $\forall k\geq k_{1}$, $d(x_{y}, s_{k})< \delta$. If $k_{2} = max\{ k_{o}, k_{1} \}$, then $\forall k > k_{2}$, $s_{k} \in B(x_{y}, \delta) \subset B(x_{y}, \varepsilon)$ and $s_{k} \in D$, $s_{k} \neq x_{y}$, therefore $\forall k > k_{2}$, $s_{k} \in D \cap B(x_{y}, \delta) \backslash \{ x_{y} \} \neq \emptyset$.\\

Thus, in any case $x_{y}$ is an accumulation point. Then by lemma \ref{lemma:eqAccumulationPointStability}, the conclusion follows. QED.

\section{Discussion}

Machine Learning models and their respective implementations include some uncertainties due to singularity regions. As such, the abstraction provided in this paper named classifier (Definition \ref{def:classifier}) is indeed a rough approximation of ML models. Nonetheless, the results obtained mostly rely upon properties of the domain of the ML model and not in the classifier itself. An important claim in this approach is that the conditions for ML stability depend, in a first place, upon certain topological and measurable properties of the space, namely density, accumulation and openness. Overall, open and bounded sets are suitable candidates sets for stable classification. As it is shown, if such topological/metric properties are not ensured, classifiers cannot ensure a stable operation. This approach shall be leveraged, by adapting the notion of classifier so as to consider the uncertainty of classifiers. This is left as a future work. The equivalence between stability and the so named accumulation series is intended to facilitate the specification of algorithms to test (absence of) stability. Whereas accumulation points and the density of sets are simple notions, they can be hard to verify on complex, high dimension domains. Then the notion of accumulation series seeks for a discrete solution, more adapted to finite/limited precision of computers. A description of such discrete algorithms was barely sketched but not formally defined.

\section{Conclusions}

This work aims to provide further understanding regarding foundational aspects of Machine Learning models. The first part of this paper introduces a definition for stability, a property that appears fundamental for the proper operation of classifiers which are defined as a function that abstracts the uncertainty of real Machine Learning models. Secondly, some limits of classifiers were explained which appear whenever one or more classification sets are dense. Indeed, it was proved that the stability of classifiers demands the absence of dense sets in the domain of classification (also called Operational Design Domain) (Lemmas \ref{lemma:densesets}, \ref{lemma:genClassifierDenseSets}). Some basic but representative examples were provided to illustrate such limitations. The last part of this work provides alternatives to prove stability in the form of equivalences: Lemma \ref{lemma:eqAccumulationPointStability} provides conditions for equivalence between stability and accumulation points, then, Lemma \ref{lemma:eqAccumPointsAndSeries} proves the equivalence between accumulation points and so named accumulation series, also introduced in this paper. The resulting equivalence between stability and accumulation series provides the possibility to define finite, discrete algorithms to prove stability. The definition of referred algorithms is not included in this paper and will be addressed as a continuation of this work.

\bibliographystyle{splncs04}
\bibliography{stabilitymath.bib}

\begin{thebibliography}{10}
\providecommand{\url}[1]{\texttt{#1}}
\providecommand{\urlprefix}{URL }
\providecommand{\doi}[1]{https://doi.org/#1}

\bibitem{Confiance.AIProject}
{Confiance.AI Project}: {The Confiance.AI Project} (2024),
  \url{https://www.confiance.ai/}

\bibitem{DEELProject}
{DEEL Project}: {Dependable, Certifiable \& Explainable Artificial Intelligence
  for Critical Systems} (2024), \url{https://www.deel.ai/}

\bibitem{EUAIAct}
{European Commission}: {Laying Down Harmonised Rules on Artificial Intelligence
  (Artificial Intelligence Act) and Amending Certain Union Legislative Acts}
  (2024),
  \url{https://eur-lex.europa.eu/legal-content/EN/TXT/?uri=CELEX:52021PC0206}

\bibitem{FischerAIsystemengineering}
Fischer, L., Ehrlinger, L., Geist, V., Ramler, R., Sobieczky, F., Zellinger,
  W., Brunner, D., Kumar, M., Moser, B.: Ai system engineering—key challenges
  and lessons learned. Machine Learning and Knowledge Extraction  \textbf{3},
  56--83 (12 2020). \doi{10.3390/make3010004}

\bibitem{Habbal2024}
Habbal, A., Ali, M.K., Abuzaraida, M.A.: Artificial intelligence trust, risk
  and security management (ai trism): Frameworks, applications, challenges and
  future research directions. Expert Systems with Applications  \textbf{240},
  122442 (2024)

\bibitem{MeasureTheoryHalmos}
Halmos, P.R.: Measure Theory. Springer New York, NY (1974),
  \url{https://link.springer.com/book/10.1007/978-1-4684-9440-2}

\bibitem{FoundationsAIHarel2020}
Harel, D., Marron, A., Sifakis, J.: Autonomics: In search of a foundation for
  next-generation autonomous systems. Proceedings of the National Academy of
  Sciences  \textbf{117}(30),  17491–17498 (jul 2020)

\bibitem{FunctionalAnalysisLang}
Lang, S.: Real and Functional Analysis. Springer New York, NY (1993),
  \url{https://link.springer.com/book/10.1007/978-1-4612-0897-6}

\bibitem{TopologicalManifoldsJohnLee}
Lee, J.M.: Introduction to Topological Manifolds. Springer New York, NY (2010),
  \url{https://link.springer.com/book/10.1007/978-1-4419-7940-7}

\bibitem{piorkowski2022quantitative}
Piorkowski, D., Hind, M., Richards, J.: Quantitative ai risk assessments:
  Opportunities and challenges. arXiv preprint arXiv:2209.06317  (2022)

\bibitem{RobustnessAI2024}
Tan, W.G.Y., Wu, Z.: Robust machine learning modeling for predictive control
  using lipschitz-constrained neural networks. Computers \& Chemical
  Engineering  \textbf{180},  108466 (2024)

\end{thebibliography}
\end{document}